\documentclass[letterpaper, 10 pt, conference]{ieeeconf}  % Comment this line out if you need a4paper

\IEEEoverridecommandlockouts                              % This command is only needed if 

\usepackage{graphicx} 
\graphicspath{ {./Images/} }
\usepackage{subcaption}

\usepackage{verbatim} 
\usepackage{changepage}
\usepackage{tikz}
\usepackage{bigstrut}
\usepackage{multirow}
\usepackage[T1]{fontenc}

\usepackage{algorithm}
\usepackage[noend]{algpseudocode}
\usepackage{amsmath}

\usepackage[export]{adjustbox}

\title{\LARGE \bf
Multi-Vehicle Trajectory Optimisation On Road Networks
}
\author{Philip Gun,$^{1,2}$ Andrew Hill$^{1,3}$, and Robin Vujanic$^{1,4}$
\thanks{*This work was supported by Rio Tinto}
\thanks{$^{1}$Philip Gun, Andrew Hill and Robin Vujanic are with the Rio Tinto Centre for Mining Automation, Australian Centre for Field Robotics, Faculty of Engineering and IT, University of Sydney, Sydney, Australia}
\thanks{$^{2,3,4}$\tt{\{pgun,a.hill,robin\}@acfr.usyd.edu.au}}
}

\begin{document}

\maketitle
\thispagestyle{empty}
\pagestyle{empty}

%%%%%%%%%%%%%%%%%%%%%%%%%%%%%%%%%%%%%%%%%%%%%%%%%%%%%%%%%%%%%%%%%%%%%%%%%%%%%%%%%%%%%
\begin{abstract} \label{sec:abstract}
This paper addresses the problem of planning time-optimal trajectories for multiple cooperative agents along specified paths through a static road network. Vehicle interactions at intersections create non-trivial decisions, with complex flow-on effects for subsequent interactions. A globally optimal, minimum time trajectory is found for all vehicles using Mixed Integer Linear Programming (MILP). Computational performance is improved by minimising binary variables using iteratively applied targeted collision constraints, and efficient goal constraints. Simulation results in an open-pit mining scenario compare the proposed method against a fast heuristic method and a reactive approach based on site practices. The heuristic is found to scale better with problem size while the MILP is able to avoid local minima.

\end{abstract}
%%%%%%%%%%%%%%%%%%%%%%%%%%%%%%%%%%%%%%%%%%%%%%%%%%%%%%%%%%%%%%%%%%%%%%%%%%%%%%%%%%%%%
%%%%%%%%%%%%%%%%%%%%%%%%%%%%%%%%%%%%%%%%%%%%%%%%%%%%%%%%%%%%%%%%%%%%%%%%%%%%%%%%%%%%%
\section{INTRODUCTION} \label{sec:intro}
%%%%%%%%%%%%%%%%%%%%%%%%%%%%%%%%%%%%%%%%%%%%%%%%%%%%%%%%%%%%%%%%%%%%%%%%%%%%%%%%%%%%%
%%%%%%%%%%%%%%%%%%%%%%%%%%%%%%%%%%%%%%%%%%%%%%%%%%%%%%%%%%%%%%%%%%%%%%%%%%%%%%%%%%%%%

\begin{figure}[ht]
    \fbox{\includegraphics[width=3.8cm, height=8.3cm, angle=90]{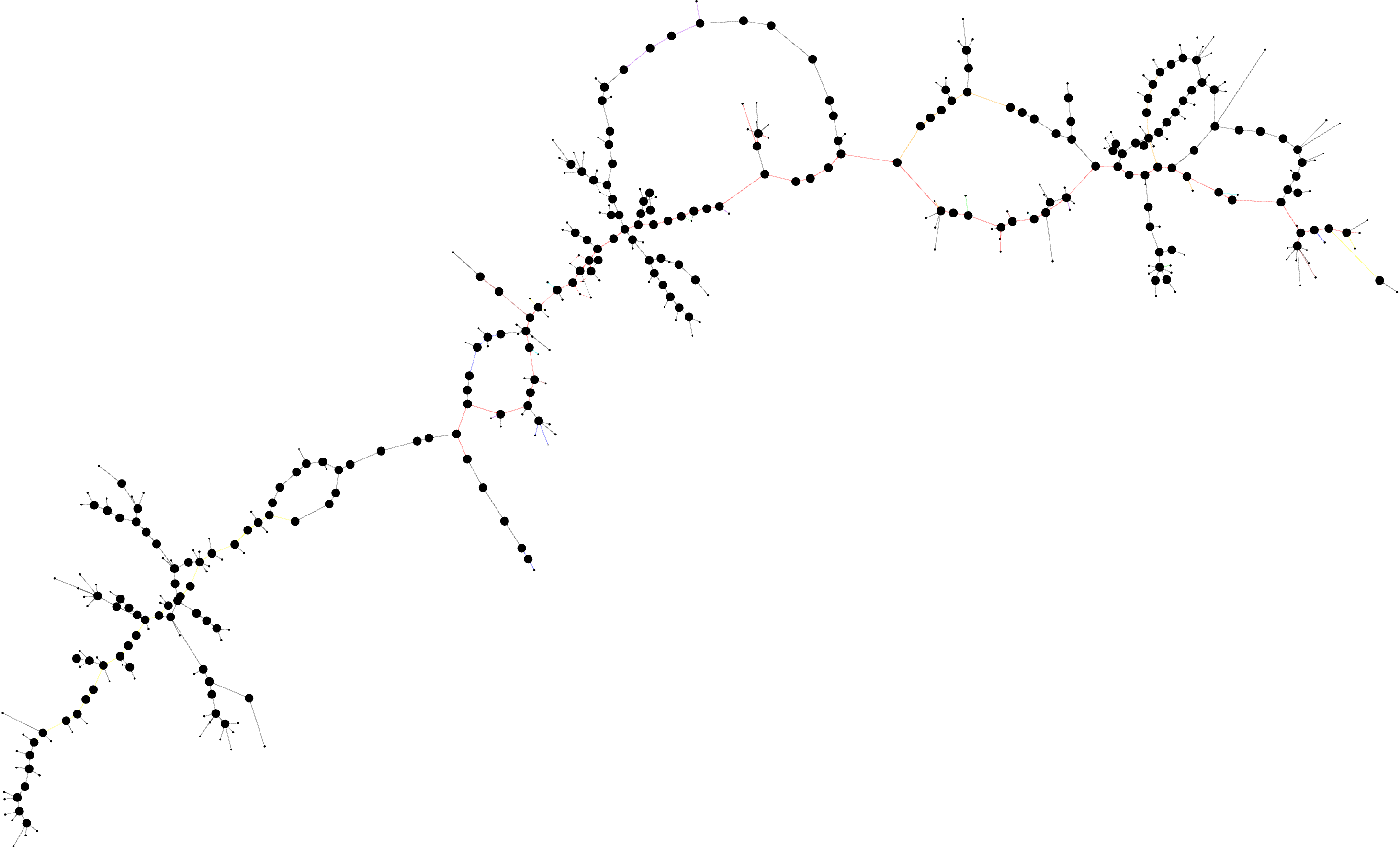}}
    \caption{Graph of a mine road network. Coloured edges represent vehicle paths defined by task assignments}
    \label{fig:mine}
    \vspace*{-\baselineskip}
\end{figure}

The work in this paper is motivated by surface mining operations. Mine road networks connect many points of interest where raw material is extracted, stored and processed. Multiple haul trucks transport material between sources and sinks. Fig. \ref{fig:mine} shows a topologically equivalent graph of a mine road network (geometrically obfuscated for commercial reasons), with many sinks, sources, and intersections where vehicles may interact.

The overall goal in mining is to maximise material extraction. To ensure this, trucks attempt to reach their destinations in minimal time and simply travel at maximum speed, with little consideration of how their actions affect each other. 

The aim of this paper is to develop a multi-vehicle trajectory planner that minimises the fleet's total traversal time. A major part of this involves making local decisions in response to interactions between vehicles, which can have complex flow-on effects on multiple other vehicles, i.e. interdependencies/coupling between vehicles.

Globally optimal methods often only solve small problems within practical computation times \cite{RN791}. The problem is challenging because resolving interactions amounts to large combinatorial problems. The number of solutions increases exponentially with the number of interactions. Interactions at intersections capture these aspects, and are the focus of this paper. Interactions from sharing road sections greatly increase the complexity of the problem. To solve optimally they require additional methods out of the scope of this paper. This paper considers only intersection interactions. 

Mixed Integer Linear Programming (MILP) is used to incorporate the entire trajectory optimisation problem, and is solved with Gurobi \cite{RN1172}. The main contribution is an extension of an iterative MILP planning method, applied to multiple cooperative vehicles on road networks. An improvement to the method is also presented that reduces iterations and computation time.

A distance based objective function (OF) that avoids binary variables is formulated and analysed. Lower and upper bounds on goal times are found allowing more efficient goal constraints and fewer binary variables with a time based OF. 

Additionally, a faster locally optimal heuristic algorithm is presented, as well as a planner to imitate the behaviour of real mine trucks. These typically react to imminent interactions at intersections as they are encountered during operation. Simulated experiments test and compare the reactive, heuristic, and MILP methods.

%%%%%%%%%%%%%%%%%%%%%%%%%%%%%%%%%%%%%%%%%%%%%%%%%%%%%%%%%%%%%%%%%%%%%%%%%%%%%%%%%%%%%
%%%%%%%%%%%%%%%%%%%%%%%%%%%%%%%%%%%%%%%%%%%%%%%%%%%%%%%%%%%%%%%%%%%%%%%%%%%%%%%%%%%%%
\section{LITERATURE REVIEW} \label{sec:litRev}
%%%%%%%%%%%%%%%%%%%%%%%%%%%%%%%%%%%%%%%%%%%%%%%%%%%%%%%%%%%%%%%%%%%%%%%%%%%%%%%%%%%%%
%%%%%%%%%%%%%%%%%%%%%%%%%%%%%%%%%%%%%%%%%%%%%%%%%%%%%%%%%%%%%%%%%%%%%%%%%%%%%%%%%%%%%

MILP can be used for single vehicle trajectory planning \cite{RN883}, \cite{RN865}, \cite{RN930}, or for multiple cooperative vehicles \cite{RN784}, \cite{RN938}, \cite{RN950}. Dynamics are modelled as discrete time systems solved as MILPs when minimising traversal time. Big-M constraints force vehicles to stay out of static obstacle areas at each time step. Binary variables select which side of obstacles vehicles pass. Moving obstacles with known trajectories are similarly avoided by changing the avoidance area at every time step \cite{RN783}. A similar formulation to moving obstacles extends to multiple vehicles avoiding inter-vehicle collisions. Trajectories are optimised as part of the MILP, rather than predefined. This greatly increases the number of binary variables and can easily result in impractical solve time. Binary variables select the order of travel between a pair of vehicles through a common area.

\cite{RN945} presents an iterative MILP method to decrease trajectory planning time. The first iteration relaxes all avoidance constraints. Obstacle collisions are identified in the resulting trajectory, defining times to apply avoidance constraints in the next iteration. This process repeats until the trajectory is collision free. The work in \cite{RN945} is only applied to individual vehicles in a static 2D environment.

\cite{RN892}, \cite{RN893}, \cite{RN870}, \cite{RN895} plan multi-vehicle trajectories on road networks. They formulate two MILPs, with approximations that remove feasibility and optimality. The first assumes unbounded acceleration. The second constrains velocity to its maximum value at all road segment edges. They suffer long computation times, scaling poorly with vehicle interactions.

\cite{RN792}, \cite{RN791}, \cite{RN789} use Branch and Bound (BNB) to select vehicle travel order through shared areas as a sequence of decisions. A separate method plans paths adhering to vehicle orderings, which initialises an an interior-point solver. A downside of the multi-vehicle method (\cite{RN791}) is the remaining possibility of collision after optimisation.

Prioritised planners define a sequence of vehicles in order of relative priority \cite{RN1162}, \cite{RN1161}, \cite{RN1163}, \cite{RN898}. Lower priority vehicles "give way". These methods are globally suboptimal because trajectories are planned one at a time, and the combined search space of all vehicles is not considered. 

Some approaches use a hierarchy of complementary planners \cite{RN980}, \cite{RN1096}, \cite{RN1028}, \cite{RN905}, \cite{RN1155}. A high level approximate solution to the global problem attempts to avoid local minima. The resulting solution is used as a reference by low level methods, which consider accurate vehicle dynamics and ensure safety. Since the high level method approximates the environment, it is generally not globally optimal at the resolution of the low level. Drawbacks of methods like \cite{RN1155} is the requirement for a priority sequence.

Reactive methods avoid collisions during execution by responding to the physical presence of agents \cite{RN1141}, \cite{RN889}, \cite{RN1145}. Solving smaller local problems lowers computational time, but there is typically no global optimality or feasibility.

\cite{RN832} apply queue theory to coordinate traffic at an intersection. \cite{RN1168} find the optimal solution to an approximation of the intersection problem with convex optimisation. \cite{RN1169} use a protocol for vehicles approaching an intersection to reserve the time-space area they will occupy during their traversal. These methods only consider individual intersections rather than an entire road network. 

%%%%%%%%%%%%%%%%%%%%%%%%%%%%%%%%%%%%%%%%%%%%%%%%%%%%%%%%%%%%%%%%%%%%%%%%%%%%%%%%%%%%%
%%%%%%%%%%%%%%%%%%%%%%%%%%%%%%%%%%%%%%%%%%%%%%%%%%%%%%%%%%%%%%%%%%%%%%%%%%%%%%%%%%%%%
\section{PROBLEM OVERVIEW} \label{sec:problem}
%%%%%%%%%%%%%%%%%%%%%%%%%%%%%%%%%%%%%%%%%%%%%%%%%%%%%%%%%%%%%%%%%%%%%%%%%%%%%%%%%%%%%
%%%%%%%%%%%%%%%%%%%%%%%%%%%%%%%%%%%%%%%%%%%%%%%%%%%%%%%%%%%%%%%%%%%%%%%%%%%%%%%%%%%%%

All vehicles travel along roads of a connected network modelled as a graph $G=(V,E)$, with vertices representing sinks, sources, and intersections. Each vehicle \(i\) is given a task that defines start and goal locations in $G$. Task assignments are assumed given by another method. Shortest paths are found with graph search methods like Dijkstra. 

Each vehicle's path \(P_{i}\) consists of a sequence of connected vertices \(P_{i}=\left\{v|v\in V\right\}\), which also defines a sequence of edges to travel \(\left\{e\vert e\in E\right\}\). Each road segment represented by an edge has an associated length. Length of a path is the sum of its edge lengths. The position \(x(t)\) of a vehicle at time \(t\) is measured along one dimension, relative to its given path. The vehicle travels from \(x(0)=0\) to \(x(T)=x_f\), with \(T\) being the arrival time at the goal.

A double integrator model is used for vehicle dynamics. Velocity is \(v(t)=\dot{x}(t)\in[0,v_{max}]\). Control input is acceleration, \(u(t)=\ddot{x}(t)\in[a_{min},a_{max}]\). Velocity is assumed to be positive as the aim is to minimise traversal time. Initial and final velocities \(v_{0}=v(0), v_{f}=v(T)\) are defined by tasks.

Once all vehicles have paths, their trajectories are calculated. The major complication studied in this paper concerns resolving vehicle interactions at intersections when their paths cross. If a pair of vehicle trajectories overlap in time when crossing an intersection, those trajectories will result in collision, and the interaction is labelled \emph{active}. Otherwise, it is \emph{inactive}. To ensure safety, we therefore wish to obtain trajectories that only have inactive interactions. The objective is to minimise the aggregated traversal time of all vehicles.

%%%%%%%%%%%%%%%%%%%%%%%%%%%%%%%%%%%%%%%%%%%%%%%%%%%%%%%%%%%%%%%%%%%%%%%%%%%%%%%%%%%%%
%%%%%%%%%%%%%%%%%%%%%%%%%%%%%%%%%%%%%%%%%%%%%%%%%%%%%%%%%%%%%%%%%%%%%%%%%%%%%%%%%%%%%
\section{MILP} \label{sec:MILP}
%%%%%%%%%%%%%%%%%%%%%%%%%%%%%%%%%%%%%%%%%%%%%%%%%%%%%%%%%%%%%%%%%%%%%%%%%%%%%%%%%%%%%
%%%%%%%%%%%%%%%%%%%%%%%%%%%%%%%%%%%%%%%%%%%%%%%%%%%%%%%%%%%%%%%%%%%%%%%%%%%%%%%%%%%%%

%%%%%%%%%%%%%%%%%%%%%%%%%%%%%%%%%%%%%%%%%%%%%%%%%%%%%%%%%%%%%%%%%%%%%%%%%%%%%%%%%%%%%
\subsection{Multi-Vehicle MILP} \label{sec:mvMILP}
%%%%%%%%%%%%%%%%%%%%%%%%%%%%%%%%%%%%%%%%%%%%%%%%%%%%%%%%%%%%%%%%%%%%%%%%%%%%%%%%%%%%%

This section summarises the MILP model developed in this paper. In contrast to [32], [6], [7], [8], whose multi-vehicle models involve a 2D/3D environment with collisions possible throughout the workspace, in this paper interactions are limited to vehicle-pairs whose paths cross, and only within intersection areas. The result is a smaller model and faster optimisation.

We utilise a discrete time model for the double integrator, with sampling rate \(1/\Delta t\), in which the control signal \(u_k\) is assumed to be constant throughout each step. The resulting model is given by:
\begin{equation} \label{eq:dynamics}
    \begin{split}
        v_{k}=v_{k-1}+u_{k} \cdot  \Delta t \\
        x_{k}=x_{k-1}+\frac{ \left( v_{k}+v_{k-1} \right) }{2} \cdot  \Delta t
    \end{split}
\end{equation}
\(K\) is defined to be a high estimate of the goal time. Binary variables \(b_{k}\) for each step determine goal arrival time. A constraint ensures that exactly one \(b_{k}\) equals to 1: \(\sum _{k=1}^{K}b_{k}=1\). The associated step \(k\) is used as the traversal time. Big-M constraints at each \(k\) ensure the vehicle has velocity \(v_f\) at position \(x_{f}\) when \(b_{k}=1\).
\begin{equation} \label{eq:finalX}
    \begin{split}
        x_{k} \leq x_{f}+ \left( 1-b_{k} \right) M \\
        x_{k} \geq x_{f}- \left( 1-b_{k} \right) M \\
        v_{k} \leq v_{f}+ \left( 1-b_{k} \right) M \\
        v_{k} \geq v_{f}- \left( 1-b_{k} \right) M \\
        % \sum _{k=1}^{K}b_{k}=1
    \end{split}
\end{equation}
Where M is a large positive number.

The extension to multiple vehicles involves adding constraints to force all interactions to be inactive. Let vehicles \(i\) and \(j\) overlap a shared intersection at spatial intervals \([x_{is},x_{ie}]\) and \([x_{js},x_{je}]\) along their respective paths. At each step \(k\), four binary variables, \(c_{i,j,k,1},\ldots,c_{i,j,k,4}\), and four Big-M constraints are defined:
\begin{equation} \label{eq:avoid}
    \begin{split}
        x_{i,k} \leq x_{is}+Mc_{i,j,k,1} \\
        x_{i,k} \geq x_{ie}-Mc_{i,j,k,2} \\
        x_{j,k} \leq x_{js}+Mc_{i,j,k,3} \\
        x_{j,k} \geq x_{je}-Mc_{i,j,k,4} \\
    \end{split}
\end{equation}
When \(c_{i,j,k,l}=1\), the associated constraint is relaxed. When \(c_{i,j,k,l}=0\), the constraint is enforced, and one vehicle must be out of the intersection. The following constraint ensures that at least one avoidance constraint is enforced at step \(k\):
\begin{equation} \label{eq:binaryAvoid}
    \sum _{l=1}^{4}c_{i,j,k,l} \leq 3
\end{equation}
The OF minimises the sum of traversal times of all \(I\) vehicles:
\begin{equation} \label{eq:objMulti}
    J= \min_{b_{i,k}} \sum _{i=1}^{I} \sum _{k=1}^{K} \Delta t \cdot k \cdot b_{i,k}   
\end{equation}
The \(c_{i,j,k,l}\) variables determine the vehicle travel order through intersections, and when they bypass each other. The heuristic algorithm in Section \ref{sec:heuristic} constrains both aspects with waypoint constraints, one vehicle at a time. Similarly to prioritised approaches \cite{RN1162}, \cite{RN1161}, \cite{RN898}, this results in local optimality. The advantage of the MILP method is it considers different travel orders and allows both vehicles in an interaction to adjust trajectories to resolve an interaction, resulting in global optimality.

%%%%%%%%%%%%%%%%%%%%%%%%%%%%%%%%%%%%%%%%%%%%%%%%%%%%%%%%%%%%%%%%%%%%%%%%%%%%%%%%%%%%%
\subsection{Efficient Goal Constraints} \label{sec:goals}
%%%%%%%%%%%%%%%%%%%%%%%%%%%%%%%%%%%%%%%%%%%%%%%%%%%%%%%%%%%%%%%%%%%%%%%%%%%%%%%%%%%%%
A reduction in the quantity of integer variables in an MILP generally decreases optimisation time \cite{RN938}. The solution performance can be improved by reducing the search space, by imposing lower and upper bound (LB/UB) constraints on the goal arrival time of vehicles. Let $\underline{t}_i$ be the goal time of truck $i$ while neglecting all interactions, i.e., constraints \ref{eq:avoid}-\ref{eq:binaryAvoid}. Therefore: \(b_{i,k} = 0, \forall k \leq \underline{t}_i\).

Using the heuristic procedure of Section \ref{sec:heuristic} a feasible trajectory for all vehicles can be efficiently computed, which determines the goal times $\overline{t}_i$ of all vehicles $i \in I$. A "delay" is then computed for each vehicle as $\overline{t}_i - \underline{t}_i$. An optimal solution cannot contain a truck arrival time that is later than $\underline{t}_i + \sum_{i' \in I} (\overline{t}_i' - \underline{t}_i')$. For time steps $k$ after this time: $b_{i,k} = 0$ are applied.

These techniques are applied for all approaches presented in Section \ref{sec:targeted}, as well for all the numerical results in experiments in Section \ref{sec:results}, unless otherwise stated.

%%%%%%%%%%%%%%%%%%%%%%%%%%%%%%%%%%%%%%%%%%%%%%%%%%%%%%%%%%%%%%%%%%%%%%%%%%%%%%%%%%%%%
\subsection{Distance-based objective function} \label{sec:distanceOF}
%%%%%%%%%%%%%%%%%%%%%%%%%%%%%%%%%%%%%%%%%%%%%%%%%%%%%%%%%%%%%%%%%%%%%%%%%%%%%%%%%%%%%

This section eliminates the binary variables that define when vehicles reach their goals, which generally reduces computation time. An alternative OF is presented that sums the absolute distance-to-goal values:
\begin{equation} \label{eq:objDist}
    J = \min_{x_{i,k}} \sum _{i=1}^{I} \sum _{k=1}^{K} \vert x_{i,f}-x_{i,k} \vert
\end{equation}
The absolute terms in the OF are modified to make them linear. Details in \cite{RN1174}.

A limitation with OF \ref{eq:objDist} is that final velocity constraints can no longer be enforced: Only zero final velocity problems are suitable. Another is undesirable behaviour in the multi-vehicle case. The OF includes the distance-to-goal throughout the entire trajectory. Locations far from the goal early in the trajectory are penalised, and therefore so are trajectories with low velocities early on. Each vehicle's behaviour becomes greedier in attempting to reach the goal. This can be a problem when low velocity early on allows a better solution for the vehicle fleet. 

A workaround is to eliminate unnecessary distance-to-goal penalties for particular time steps by using the LB/UB on goal times in section \ref{sec:goals}. The smaller range of steps to which the penalty is applied results in more direct optimisation of goal time and better behaved trajectories.

%%%%%%%%%%%%%%%%%%%%%%%%%%%%%%%%%%%%%%%%%%%%%%%%%%%%%%%%%%%%%%%%%%%%%%%%%%%%%%%%%%%%%
%%%%%%%%%%%%%%%%%%%%%%%%%%%%%%%%%%%%%%%%%%%%%%%%%%%%%%%%%%%%%%%%%%%%%%%%%%%%%%%%%%%%%
\section{ITERATIVE SOLVERS} \label{sec:solvers}
%%%%%%%%%%%%%%%%%%%%%%%%%%%%%%%%%%%%%%%%%%%%%%%%%%%%%%%%%%%%%%%%%%%%%%%%%%%%%%%%%%%%%
%%%%%%%%%%%%%%%%%%%%%%%%%%%%%%%%%%%%%%%%%%%%%%%%%%%%%%%%%%%%%%%%%%%%%%%%%%%%%%%%%%%%%

%%%%%%%%%%%%%%%%%%%%%%%%%%%%%%%%%%%%%%%%%%%%%%%%%%%%%%%%%%%%%%%%%%%%%%%%%%%%%%%%%%%%%
% \subsection{Targeted interaction constraints} \label{sec:targeted}
\subsection{MILP With Targeted Interaction Constraints} \label{sec:targeted}
%%%%%%%%%%%%%%%%%%%%%%%%%%%%%%%%%%%%%%%%%%%%%%%%%%%%%%%%%%%%%%%%%%%%%%%%%%%%%%%%%%%%%

As shown in Section \ref{sec:results}, the largest effect on the optimisation time of the MILP is from binary variables in the interaction constraints. Reducing them is the subject of this section. For the majority of a vehicle's trajectory, it will likely not be involved in an interaction. Rather than the typical approach of applying interaction constraints at all time steps, they can be reduced by only applying them to steps they are needed.

Algorithm \ref{alg:MILP} adds interaction constraints to the model in a lazy fashion, similarly to \cite{RN945}. It begins by creating a relaxed MILP model on line \ref{alg:relaxedModel} without avoidance constraints. Active interactions are identified (line \ref{alg:active}), and avoidance constraints are added for each (line \ref{alg:addConstraints}). The model is re-optimised on line \ref{alg:optimise2} and the process is repeated until no active interactions remain (line \ref{alg:MILPwhile}). 

Line \ref{alg:interactionTimes} selects time steps to add avoidance constraints. \cite{RN945} applies one constraint at the midpoint of an interaction, potentially requiring many iterations for a feasible solution. Instead, a feature of the road network can be exploited, that vehicles must travel across known intersection regions. Avoidance constraints are added at every step in the time-interval between the first vehicle entering the intersection, and the last vehicle exiting. The usual result is only one additional iteration. As shown in section \ref{sec:results}, the interaction interval approach reduces solve times from the midpoint method of \cite{RN945}. Optimal solutions returned by Algorithm \ref{alg:MILP} are also optimal to the monolithic MILP of section \ref{sec:mvMILP}.

\begin{algorithm}
    \caption{Iterative lazy interaction constraint MILP}\label{alg:MILP}
    \begin{algorithmic}[1]
        \State \textbf{Input}: Network $G(V,E)$, Vehicles with tasks $A$
        \State \textbf{Output}: Multiple vehicle trajectories $S$
        \State \(M \gets \mathrm{RelaxedMultiMILP}(G,A)\) \label{alg:relaxedModel}
        \State \(S \gets \mathrm{Optimise(M)}\)
        \State $N \gets \mathrm{ActiveInteractions}(S)$ \label{alg:active}
        \While{$N\ne0$} \label{alg:MILPwhile} \Comment{While active interactions remain} 
            \ForAll{\(n \in N\)}
                \State \( (t_1,...,t_n) \gets \mathrm{InteractionTimes}(n)\) \label{alg:interactionTimes}
                \State \(M \gets \mathrm{AddAvoidanceCons}(M,n,t_1,...,t_n)\) \label{alg:addConstraints}
            \EndFor
            \State \(S \gets \mathrm{Optimise}(M)\) \label{alg:optimise2}
            \State $N \gets \mathrm{ActiveInteractions}(S) $
        \EndWhile 
        \State \textbf{return} $S$
    \end{algorithmic}
\end{algorithm}

%%%%%%%%%%%%%%%%%%%%%%%%%%%%%%%%%%%%%%%%%%%%%%%%%%%%%%%%%%%%%%%%%%%%%%%%%%%%%%%%%%%%%
%%%%%%%%%%%%%%%%%%%%%%%%%%%%%%%%%%%%%%%%%%%%%%%%%%%%%%%%%%%%%%%%%%%%%%%%%%%%%%%%%%%%%
% \section{MULTI-VEHICLE HEURISTIC SOLVER} \label{sec:heuristic}
\subsection{Sequential Avoidance Heuristic Solver} \label{sec:heuristic}
%%%%%%%%%%%%%%%%%%%%%%%%%%%%%%%%%%%%%%%%%%%%%%%%%%%%%%%%%%%%%%%%%%%%%%%%%%%%%%%%%%%%%
%%%%%%%%%%%%%%%%%%%%%%%%%%%%%%%%%%%%%%%%%%%%%%%%%%%%%%%%%%%%%%%%%%%%%%%%%%%%%%%%%%%%%
Even with the proposed improvements of Sections \ref{sec:goals} and \ref{sec:targeted}, as the problem scales up and the number of interactions increase, MILP solve times can become impractically long. A faster alternative method is presented in iterative Algorithm \ref{alg:iterate}, which sequentially resolves active interactions.

Lines \ref{alg:relaxed} and \ref{alg:iterativeActive0} first solve the relaxed problem and find all active interactions. In each iteration, the earliest active interaction is identified (line \ref{alg:earlyInteraction}) between two vehicles. Let $i$ be the vehicle \emph{entering} the intersection first, and $j$ the last (line \ref{alg:sort}). A spatiotemporal waypoint constraint is added to the model to resolve the interaction, forcing $j$ to wait until $i$ has left the intersection. Let $\tilde{t}_{i}$ (line \ref{alg:exitTime}) be the time at which vehicle $i$ \emph{leaves} the intersection; then, the constraint added to the model is \(x_{j,k} \leq {x}_{js}\). \(k\) is the closest time step to \(\tilde{t}_i\), and \({x}_{j,s}\) is where \(j\) enters the intersection (line \ref{alg:entrancePos}).

Line \ref{alg:resolve} calls Algorithm \ref{alg:heuristic}, which calculates a new trajectory for \(j\). A a single vehicle MILP is used, adding waypoint constraints for every interaction resolution. After trajectory adjustment, active interactions are re-identified on line \ref{alg:iterativeActive}, and the process repeats until no active interactions remain. Each vehicle has a sequence of waypoints at the end.

Although not tested in this paper, an alternative is to use equality waypoint constraints: \(x_{j,k}={x}_{js}\). The idea is that the closest feasible point to an active interaction results in a close to optimal trajectory, assuming the prior trajectory solution is optimal in the relaxed problem. The advantage is the equality results in a smaller search space relative to the inequality, and therefore will take less time to optimise.

\begin{algorithm}
    \caption{Sequential Avoidance Heuristic}\label{alg:iterate}
    \begin{algorithmic}[1]
        \State \textbf{Input}: Network $G(V,E)$, Vehicles with tasks $A$
        \State \textbf{Output}: Multiple vehicle trajectories $S$
        \State Calculate initial relaxed trajectories $S$ \label{alg:relaxed}
        \State $N \gets \mathrm{ActiveInteractions}(S)$ \label{alg:iterativeActive0}
        \While{$N\ne0$} \Comment{While active interactions remain}
            \State $n_0 \gets \mathrm{EarliestInteraction}(N) $ \label{alg:earlyInteraction}
            \State $i,j \gets \mathrm{SortVehicles}(n_0) $ \Comment{Non-adj., Adjusting} \label{alg:sort}
            \State ${x}_{js} \gets \mathrm{EntrancePosition}(n_0,j) $ \label{alg:entrancePos}
            \State $\tilde{t}_{i} \gets \mathrm{DepartureTime}(n_0,i) $ \label{alg:exitTime} 
            \State $s \gets \mathrm{ResolveInteraction}(j,\tilde{t}_{i},{x}_{js}) $ \label{alg:resolve}
            \State $S[j] \gets s $ \Comment{Update solution set} \label{alg:update}
            \State $N \gets \mathrm{ActiveInteractions}(S) $ \label{alg:iterativeActive}
        \EndWhile 
        \State \textbf{return} $S$
    \end{algorithmic}
\end{algorithm}

% HEURISTIC
\begin{algorithm}
    \caption{Waypoint interaction resolver}\label{alg:heuristic}
    \begin{algorithmic}[1]
        \State \textbf{Procedure} \(\mathrm{ResolveInteraction}(j,\tilde{t}_{i},{x}_{js},Heuristic)\):
            \State \(j[WP] \gets j[WP] + \mathrm{WaypointConstraint}(\tilde{t}_{i},{x}_{js})\) \label{alg:waypoint} 
            \State \(s \gets \mathrm{SingleVehicleMILP}(j)\) \Comment{New trajectory for \(j\)} \label{alg:heurMILP}
            \State \textbf{return} $s$
    \end{algorithmic}
\end{algorithm}

%%%%%%%%%%%%%%%%%%%%%%%%%%%%%%%%%%%%%%%%%%%%%%%%%%%%%%%%%%%%%%%%%%%%%%%%%%%%%%%%%%%%%
%%%%%%%%%%%%%%%%%%%%%%%%%%%%%%%%%%%%%%%%%%%%%%%%%%%%%%%%%%%%%%%%%%%%%%%%%%%%%%%%%%%%%
% \section{REACTIVE SOLVER} \label{sec:reactive}
\subsection{Reactive interaction resolver} \label{sec:reactive}
%%%%%%%%%%%%%%%%%%%%%%%%%%%%%%%%%%%%%%%%%%%%%%%%%%%%%%%%%%%%%%%%%%%%%%%%%%%%%%%%%%%%%
%%%%%%%%%%%%%%%%%%%%%%%%%%%%%%%%%%%%%%%%%%%%%%%%%%%%%%%%%%%%%%%%%%%%%%%%%%%%%%%%%%%%%

To compare the presented MILP methods, a reactive solver was developed that approximates the control method haul trucks use on real mines. The iterative framework of Algorithm \ref{alg:iterate} is used again, with two key differences: The interaction resolver on line \ref{alg:resolve} calls Algorithm \ref{alg:react}, and minimum time trajectories (both initial and adjusted) for each vehicle are found by solving a two-point boundary value problem (TPBVP), which has a bang-off-bang control structure \cite{RN870}.

Interactions are resolved by one of the vehicle-pair "giving way" to the other based on visibility at the intersection. Active interactions are identified by an inflated virtual safety buffer, resulting in more conservative reactions to other vehicles. Line \ref{alg:stop} adjusts \(j\)'s trajectory to stop just before the intersection. Once \(i\) exits, \(j\) proceeds again at maximum speed (line \ref{alg:goal}), regardless of whether it has stopped.

% REACTIVE
\begin{algorithm}
    \caption{Reactive interaction resolver}\label{alg:react}
    \begin{algorithmic}[1]
        \State \textbf{Procedure} \(\mathrm{ResolveInteraction}(j,\tilde{t}_{i},{x}_{js},Reactive)\):
            \State $s_{decel} \gets \mathrm{StopTrajectory}(j,{x}_{js}) $ \label{alg:stop}
            \State $s_{goal} \gets \mathrm{GoalTrajectory}(s_{decel},\tilde{t}_{i}) $ \label{alg:goal} 
            \State $s \gets \mathrm{CombineTrajectory}(s,s_{decel},s_{goal}) $ 
            \State \textbf{return} $s$ 
    \end{algorithmic}
    % \vspace*{-5mm}
\end{algorithm}

%%%%%%%%%%%%%%%%%%%%%%%%%%%%%%%%%%%%%%%%%%%%%%%%%%%%%%%%%%%%%%%%%%%%%%%%%%%%%%%%%%%%%
%%%%%%%%%%%%%%%%%%%%%%%%%%%%%%%%%%%%%%%%%%%%%%%%%%%%%%%%%%%%%%%%%%%%%%%%%%%%%%%%%%%%%
\section{EXPERIMENTS AND RESULTS} \label{sec:results}
%%%%%%%%%%%%%%%%%%%%%%%%%%%%%%%%%%%%%%%%%%%%%%%%%%%%%%%%%%%%%%%%%%%%%%%%%%%%%%%%%%%%%
%%%%%%%%%%%%%%%%%%%%%%%%%%%%%%%%%%%%%%%%%%%%%%%%%%%%%%%%%%%%%%%%%%%%%%%%%%%%%%%%%%%%%

Experiments are run on simulated road networks in three types of scenarios. In all scenarios, vehicles begin simultaneously. Vehicle paths cross each other at multiple intersections, where interactions between vehicle-pairs occur.

\begin{figure}[b]
    \begin{subfigure}{0.55\columnwidth}
        \includegraphics[width=1\columnwidth]{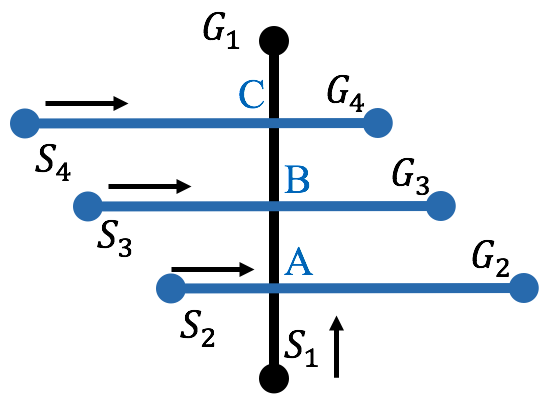}
        \caption{Toy case for heuristic solver}
        \label{fig:manual}
    \end{subfigure}
    \begin{subfigure}{0.44\columnwidth}
        \includegraphics[width=1\columnwidth]{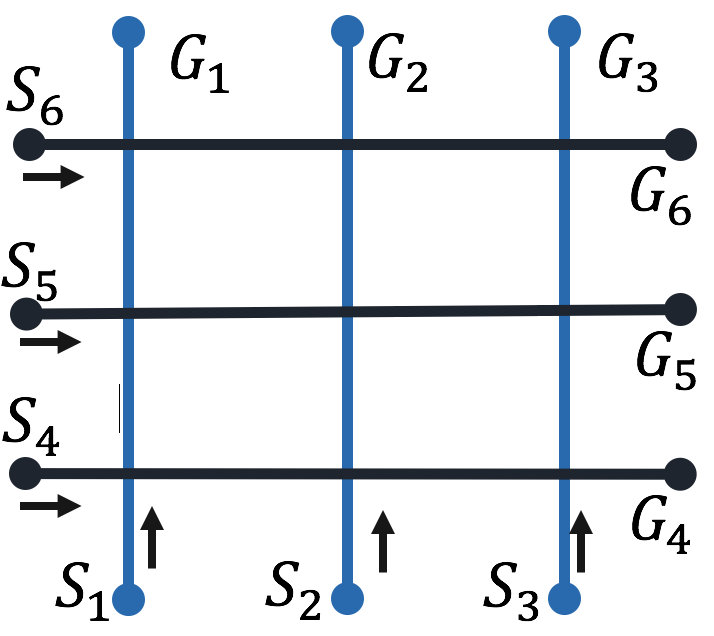} % , height=5cm
        \caption{Grid based road network}
        \label{fig:grid}
    \end{subfigure}    
    \caption{Two types of simulated road networks. \(S_i,G_i\) indicate the start and goal locations of vehicle \(i\).}
    \label{fig:networks}
    \vspace*{-\baselineskip}
\end{figure}

The first scenario type includes two toy cases, with a road network structured as in Fig. \ref{fig:manual}, to illustrate differences between the methods presented. Vehicle 1 travels a path with three consecutive intersections, each one crossed by another vehicle. In the first case, the relaxed solution results in vehicle 1 actively interacting with every other vehicle (at intersections A,B, and C), reaching each intersection just before the others. In the second case, different road lengths change vehicle behaviour. The relaxed solution results in one active interaction between vehicles 1 and 2 at intersection A, with vehicle 2 entering first. Such cases are expected to result in locally optimal but globally suboptimal solutions with greedy algorithms such as the sequential avoidance heuristic of Section \ref{sec:heuristic}. The MILP is expected to find solutions with better sequencing decisions.

The second scenario type consist of grid-like structures, which allows a consistent way to scale problems up. An example of a 3x3 grid is Fig. \ref{fig:grid}, showing vehicle start locations. Vehicles traverse the straight line paths directly in front. A similar structure is used for grid networks of other sizes. This type of network is physically symmetric. Road segment lengths between intersections are all 100m, resulting in vehicles arriving at intersections at the same time. This results in a special problem case as both travel orders to resolve an interaction are equivalent in cost and typically both must be considered.

The last scenario uses the road network shown in Fig. \ref{fig:mine}, which contains 431 nodes and 876 connecting edges. This involved creating paths in the network defined by randomly selected start and goal nodes for 24 vehicles.

All solutions were computed on a PC with Windows 7, Intel Core i7-4810MQ 2.8GHz CPU, 16GB RAM. MILPs optimised with Gurobi 8. All vehicles are identical, with models approximating mining haul trucks: Lengths $15m$; mass $200t$; coefficient of rolling resistance \(0.08\), maximum and minimum accelerations \(\pm3m/s^2\).

\begin{figure}[h]
 
    \begin{subfigure}{0.49\columnwidth}
        \includegraphics[width=1\columnwidth]{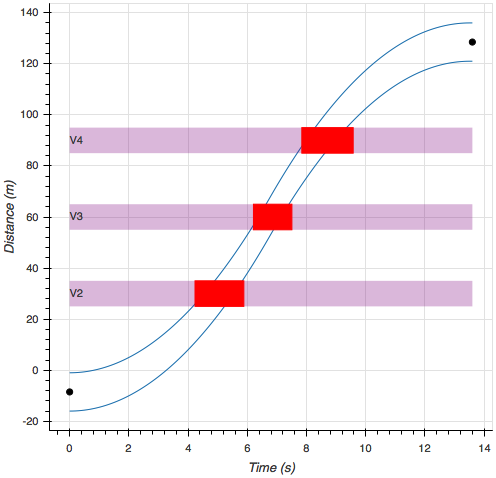} 
        \caption{Case 1 relaxed solution}
        \label{fig:manual1Start}
    \end{subfigure}
    \begin{subfigure}{0.49\columnwidth}
        \includegraphics[width=1\columnwidth]{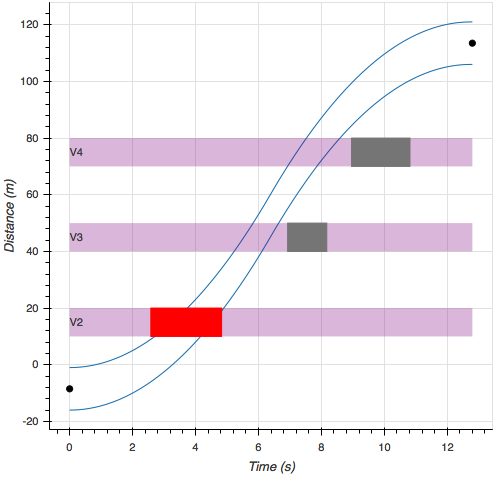}
        \caption{Case 2 relaxed solution}
        \label{fig:manual2Start}
    \end{subfigure}
    
    \begin{subfigure}{0.49\columnwidth}
        \includegraphics[width=1\columnwidth]{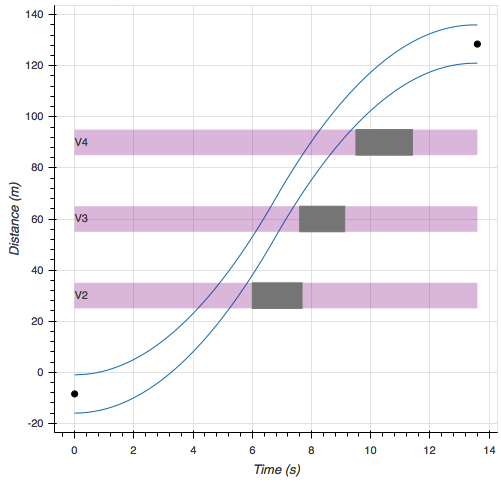} 
        \caption{Case 1 Heuristic solution}
        \label{fig:manual1Heur}
    \end{subfigure}
    \begin{subfigure}{0.49\columnwidth}
        \includegraphics[width=1\columnwidth]{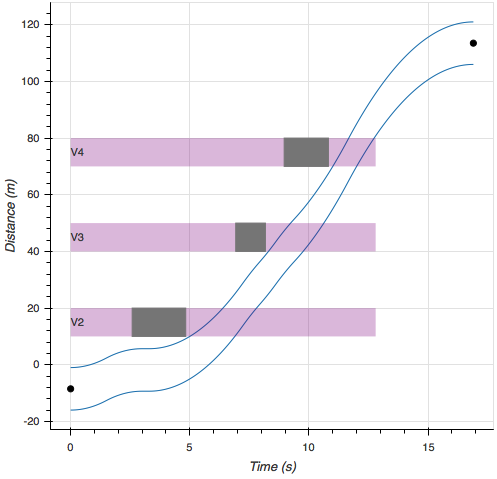}
        \caption{Case 2 Heuristic solution}
        \label{fig:manual2Heur}
    \end{subfigure}
    
    \begin{subfigure}{0.49\columnwidth}
        \includegraphics[width=1\columnwidth]{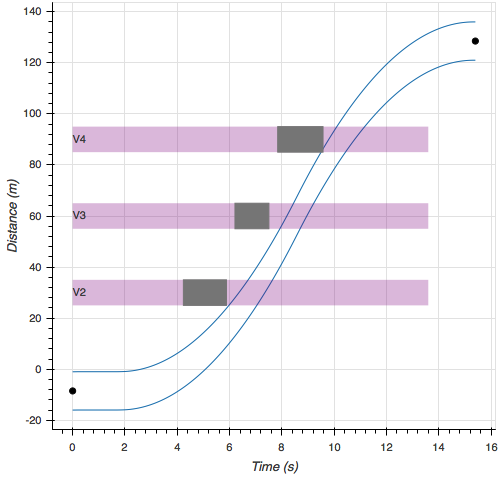} 
        \caption{Case 1 MILP solution}
        \label{fig:manual1MILP}
    \end{subfigure}
    \begin{subfigure}{0.49\columnwidth}
        \includegraphics[width=1\columnwidth]{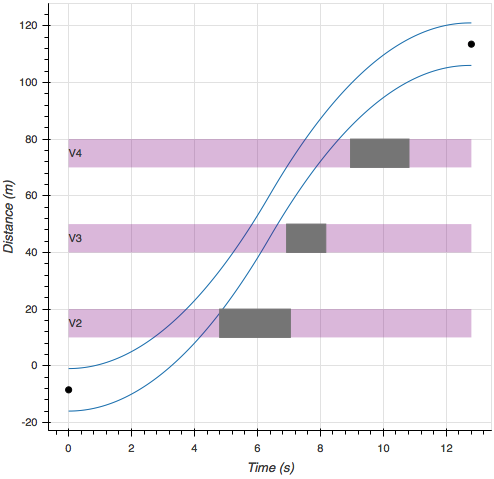}
        \caption{Case 2 MILP solution}
        \label{fig:manual2MILP}
    \end{subfigure}
    
    \caption{Trajectories of vehicle 1 in Fig. \ref{fig:manual} in two different toy cases. Blue lines show front and rear vehicle position along path, purple bars represent intersection areas. Rectangles indicate other vehicles crossing intersections: Red represents active interactions, and grey inactive.}
    \label{fig:manualTrajectories}
    \vspace*{-\baselineskip}
\end{figure}

Figs. \ref{fig:manual1Start}, \ref{fig:manual1Heur}, and \ref{fig:manual1MILP} show vehicle 1's trajectory for the first toy case in the relaxed, heuristic and MILP solutions respectively. The heuristic selects vehicle 1 to maintain its trajectory and delays all other vehicles. The total delay to the fleet is proportional to the number of vehicles. The MILP instead considers the possibility of vehicle 1 slowing down, which is sufficient for all interactions. The total delay is due to only vehicle 1's adjustment. The result is a delay time of 4s for the MILP, less than half of the heuristic's 10.9s delay.

Trajectories are shown in Figs. \ref{fig:manual2Start}, \ref{fig:manual2Heur}, and \ref{fig:manual2MILP} for the second toy case. To avoid the active interaction at intersection A, the heuristic delays vehicle 1's trajectory. This causes a new interaction at intersection B with vehicle 3. Vehicle 1 is further delayed, and a chain of delays results. The MILP instead delays vehicle 2 once rather than multiple times. The experiment results in a 2.4s delay for the MILP, and 7.1s for the heuristic. These experiments show that the MILP is able to find better solutions than the heuristic in particular problem structures.

Table \ref{tab:OFgoal} shows results of experiments comparing computation times of the time-based and distance-based OF, applied to grid networks of various size. Solving the model with distance-based OF is faster, ranging between 11\% to 24\% of the time-based OF. Table \ref{tab:OFgoal} also shows the effect of applying a narrower range of goal constraints as discussed in section \ref{sec:goals}. The narrower range results in solve times ranging between 24\% and 58\%  of the full range model.

\begin{table}[htbp]
  \centering
    \begin{tabular}{|c|c|c|c|}
    \hline
          & \multicolumn{2}{c|}{Time OF} & Distance OF \bigstrut\\
    \hline
    Grid size & Narrow range & \multicolumn{2}{c|}{Full range} \bigstrut\\
    \hline
    1     & 0.6   & 2.5   & 0.3 \bigstrut\\
    \hline
    2     & 2.7   & 9.9   & 2.0 \bigstrut\\
    \hline
    3     & 11.0  & 19.2  & 4.8 \bigstrut\\
    \hline
    \end{tabular}%
    \caption{Computation time (s) comparison of two OFs}
    \label{tab:OFgoal}%
    \vspace*{-\baselineskip}
\end{table}%

Table \ref{tab:compTime} shows the solve times of the iterative MILP methods of section \ref{sec:targeted}. The second column refers to the model with interaction avoidance constraints at all steps. Only shown up to the 4x4 size as larger problems result in long solve times. The third column refers to the iterative midpoint constraint method, which gains a computational reduction ranging between 86\% and 96\%. The fourth column refers to constraints applied to all steps during active interactions, gaining a further reduction between 18\% to 60\%. The last column refers to the heuristic of section \ref{sec:heuristic}, which is typically faster than the rest, with solve times reduction as high as 63\% over the fastest MILP method.

In terms of scaling up with problem size, the interaction-interval method scales better than the midpoint method, though not as well as the heuristic, which is approximately linear. The reactive method solve times were consistently under 10ms for all problems.

\begin{table}[htbp]
    \centering
    \begin{tabular}{|c|c|c|c|c|}
        \hline
        \multicolumn{1}{|p{1.75em}|}{Grid} & MILP  & \multicolumn{2}{c|}{Iterative MILP} & Seq. Avoid.  \bigstrut\\
        \cline{2-4}    \multicolumn{1}{|p{1.75em}|}{size} & Full range & Midpoint & Interaction interval & \multicolumn{1}{p{4.5em}|}{Heuristic} \bigstrut\\
        \hline
        1     & 2.5   & 0.34  & 0.14  & 0.15 \bigstrut\\
        \hline
        2     & 9.9   & 0.80  & 0.38  & 0.41 \bigstrut\\
        \hline
        3     & 19    & 1.5   & 0.73  & 0.58 \bigstrut\\
        \hline
        4     & 59    & 2.1   & 1.2   & 0.88 \bigstrut\\
        \hline
        6     & > 60  & 4.3   & 2.8   & 1.4 \bigstrut\\
        \hline
        8     & > 60  & 7.0   & 4.2   & 1.9 \bigstrut\\
        \hline
        10    & > 60  & 11    & 5.6   & 2.5 \bigstrut\\
        \hline
    \end{tabular}%
    \caption{Computation times (s) of various methods applied to grid networks. Full range refers to adding interaction-avoidance constraints to every time step of the MILP. The two iterative methods add targeted constraints.}
  \label{tab:compTime}
  \vspace*{-\baselineskip}
\end{table}%

\begin{figure}[t]
    \centering
    \includegraphics[width=1\columnwidth]{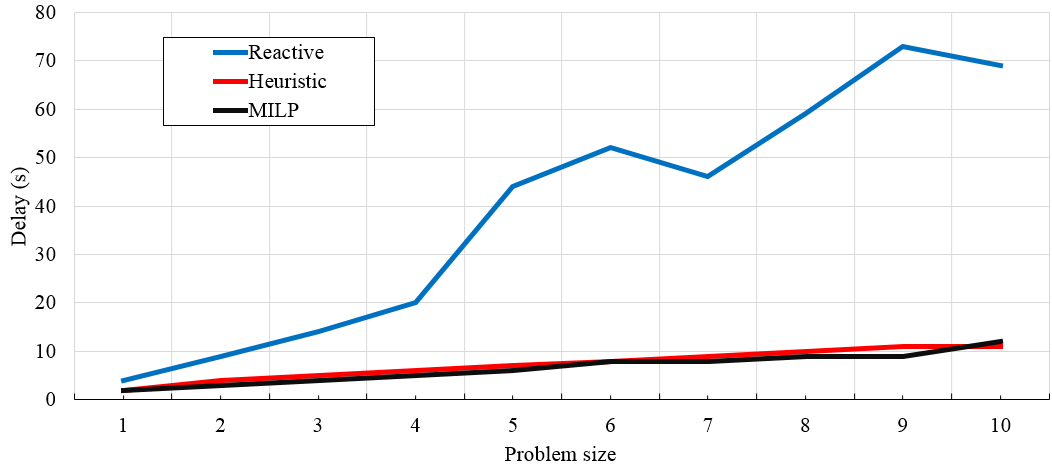}
    \caption{Delay time of the various methods as the problem size is scaled up}
    \label{Fig:delay}
    \vspace*{-\baselineskip}
\end{figure}

Fig. \ref{Fig:delay} displays the solution performance of various methods, measured as total delay time, defined in Section \ref{sec:goals}. The reactive solution consistently results in the highest total delay. The heuristic usually has a delay slightly higher than the MILP method. Only one MILP line is shown as the various MILP methods of Section \ref{sec:MILP} give the same answers, varying only in solve times.

% Execution time on top of the relaxed solution. 

The delay data shows the heuristic solver works approximately as well as the MILP optimiser in terms of solution cost, and often with lower computation time for grid structured models. This can be attributed to the optimal solution consisting of any one of the two vehicles at each active interaction giving way, which does not affect any other vehicle. In these scenarios the heuristic matches this behaviour, and complexity that the MILP can capture has no advantage.

\begin{figure}[ht]
    \begin{subfigure}{0.49\columnwidth}
        \includegraphics[width=1\columnwidth]{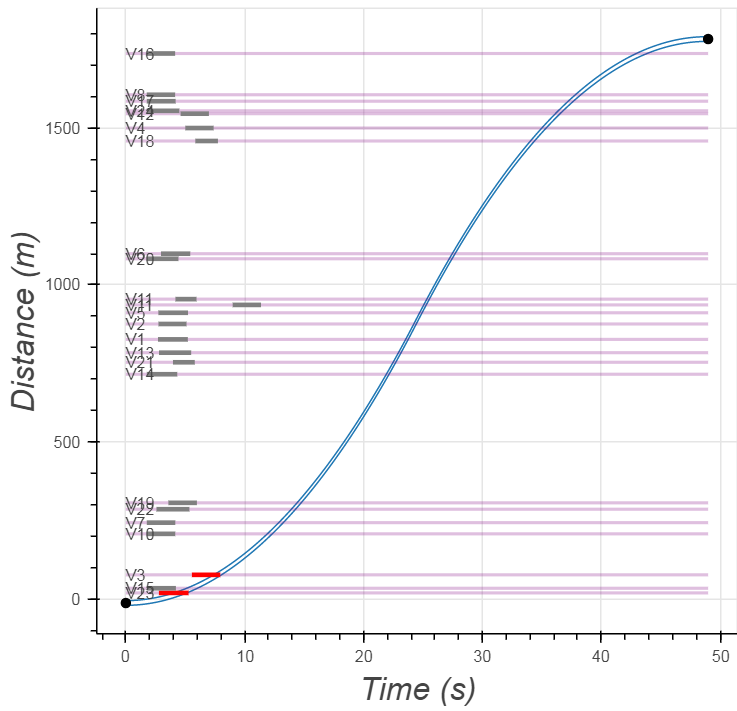}
        \caption{Vehicle 9 with long path crossing many intersections.}
        \label{fig:v9Traj}
    \end{subfigure}
    \begin{subfigure}{0.49\columnwidth}
        \includegraphics[width=1\columnwidth]{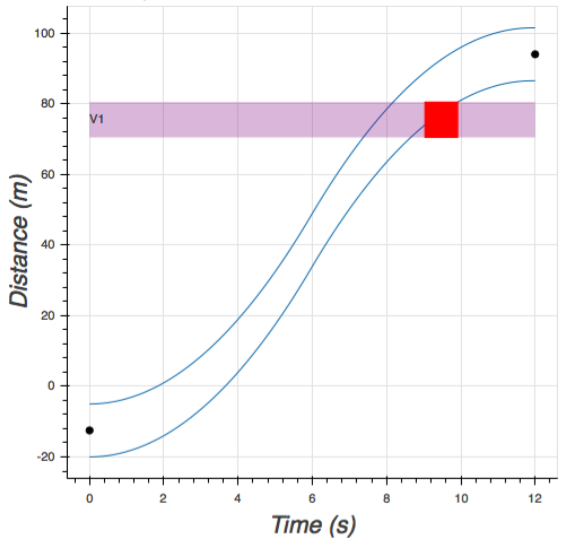} % , height=5cm
        \caption{Vehicle 23 with short path crossing one intersection.}
        \label{fig:v23Traj}
    \end{subfigure}    
    \caption{Trajectories of two interacting vehicles on the road network of Fig. \ref{fig:mine}.}
    \label{fig:mineTrajectory}
    \vspace*{-\baselineskip}
\end{figure}

Fig. \ref{fig:mineTrajectory} shows the relaxed solution trajectories of two vehicles on the network of Fig. \ref{fig:mine}. Despite 24 vehicles, only two active interactions occur. The path in Fig. \ref{fig:v9Traj} has many intersections but almost all interactions are inactive, while Fig. \ref{fig:v23Traj} shows a short path crossing one intersection. These are representative of problems that typically arise in topologically similar networks. The interaction sparsity is due to the network containing long road segments without intersections, and the vehicles' simultaneous starting times. 

Table \ref{tab:mine} summarises the results of multiple methods of another problem instance with 38 vehicles and 24 active interactions. The proposed iterative MILP method is better or equal performing than all methods and faster than all non-reactive planners. The Heuristic's lower solve time shows its superior scaling.

When there is interaction sparsity, the MILP only has a small delay time advantage over the Heuristic. For sparse problems, a local search method such as the Sequential Avoidance Heuristic is likely to achieve close to optimal solutions. As the problem is scaled up, the trade-off in computation time of the globally optimal MILP method increases.

\begin{table}[htbp]
  \centering
    \begin{tabular}{|l|c|r|}
    \hline
          & \multicolumn{1}{l|}{Comp. time (s)} & \multicolumn{1}{l|}{Delay (s)} \bigstrut\\
    \hline
    Reactive & 0.002 & 46.6 \bigstrut\\
    \hline
    Heuristic & 1.14  & 27.3 \bigstrut\\
    \hline
    Midpoint MILP & 10.3  & 21.8 \bigstrut\\
    \hline
    Interval MILP & 6.22  & 21.8 \bigstrut\\
    \hline
    \end{tabular}%
  \caption{Computation time and delay costs of the solution of a randomised problem on the network of Fig. \ref{fig:mine}}
    \label{tab:mine}%
    \vspace*{-\baselineskip}
\end{table}%

%%%%%%%%%%%%%%%%%%%%%%%%%%%%%%%%%%%%%%%%%%%%%%%%%%%%%%%%%%%%%%%%%%%%%%%%%%%%%%%%
%%%%%%%%%%%%%%%%%%%%%%%%%%%%%%%%%%%%%%%%%%%%%%%%%%%%%%%%%%%%%%%%%%%%%%%%%%%%%%%%
\section{CONCLUSION AND FUTURE WORK} \label{sec:conclusion}
%%%%%%%%%%%%%%%%%%%%%%%%%%%%%%%%%%%%%%%%%%%%%%%%%%%%%%%%%%%%%%%%%%%%%%%%%%%%%%%%
%%%%%%%%%%%%%%%%%%%%%%%%%%%%%%%%%%%%%%%%%%%%%%%%%%%%%%%%%%%%%%%%%%%%%%%%%%%%%%%%

A method to solve multi-vehicle trajectory planning problems on road networks has been presented. The approach consists of iteratively solving relaxed MILP models, adding targeted interaction-avoidance constraints in each iteration until a feasible and globally optimal solution emerges. More efficient versions of the OF and goal constraints were formulated and tested. An additional heuristic algorithm was developed with faster solve times but suboptimal solutions. Experimental results showed that the iterative MILP method reduced solve times relative to previous methods. 

The presented methods will be extended to resolving interactions along shared road segments which is expected to greatly increase the complexity of the models. To test the advantages and trade-offs of the optimal MILP method, experiments will be extended to scenarios where vehicle interactions are denser. Other future work includes optimising goal times based on machinery status and queues, and incorporating energy in optimisations.

% \addtolength{\textheight}{-5cm}   

\bibliographystyle{IEEEtran}
\bibliography{Myrefs}

\end{document}